\documentclass{article}
\usepackage{ijcai26}

\usepackage{times}
\usepackage{soul}
\usepackage{url}
\usepackage[hidelinks]{hyperref}
\usepackage[utf8]{inputenc}
\usepackage[small]{caption}
\usepackage{graphicx}
\usepackage{amsmath}
\usepackage{amsthm}
\usepackage{booktabs}
\usepackage{algorithm}
\usepackage{algorithmic}
\usepackage{placeins}
\usepackage[switch]{lineno}
\usepackage{amssymb}
\usepackage{tikz}
\usepackage{pifont} 
\usepackage{booktabs}
\usepackage{siunitx}
\usepackage{multirow}
\usepackage{xcolor} 
\usepackage{colortbl}    
\usepackage{multirow}    
\usepackage{makecell}    
\usepackage{amsmath}     
\usepackage{array}    
\usepackage{amsmath,amsfonts,amssymb}
\DeclareMathAlphabet{\mathbbold}{U}{bbold}{m}{n}
\title{Spherical Physics-informed Neural Operator with Multi-scale Coupling for Meteorological Downscaling
}

\author{
Yiqiang~Ye$^{1,}$\thanks{These authors contributed equally.}\and
Yichi~Wang$^{2,}$\footnotemark[1]\and
Jiawei~Wen$^1$\and
Jiahui~Jiang$^1$\and\\
Zhaoyu~Zhong$^1$\and
Jiangjian~Yu$^2$\and
Chunxia~Xiao$^{2,}$\thanks{Corresponding authors.}\And
Haodi~Zhang$^{1,}$\footnotemark[2]\\
\affiliations
$^1$College of Computer and Software Engineering, Shenzhen University, Shenzhen, China\\
$^2$School of Computer Science, Wuhan University, Wuhan, China\\
\emails
\{yeyiqiang2023, wenjiawei2023\}@email.szu.edu.cn,
2410103041@mails.szu.edu.cn,
857123107zzy@gmail.com,
\{whuhunter, jiangjianyu, cxxiao\}@whu.edu.cn,
hdzhang@szu.edu.cn
}

\begin{document}

\maketitle


\begin{abstract}
Meteorological downscaling is crucial for high-resolution regional climate forecasting and disaster early warning.
While neural operators have emerged as a promising paradigm for modeling complex spatiotemporal mappings, existing frameworks often struggle with spherical manifold geometric distortions, inherent atmospheric multi-scale coupling mismatches, and lack of explicit atmospheric laws. We propose the Spherical Physics-informed Neural Operator, which utilizes a Spherical Laplacian Decomposition to partition atmospheric fields into hierarchical frequency components, maintaining exact point-wise correspondence across scales. 
To evaluate these representations at arbitrary locations, we introduce a localized spherical integral operator that approximates continuous kernel transforms via geometry-aware attention. 
Dynamical consistency is further enforced by embedding differentiable constraints into the learning process. 
Extensive experiments demonstrate that our framework attains superior accuracy and zero-shot generalization across various meteorological variables and unseen queries, representing a robust and interpretable solution for global-to-regional meteorological downscaling.
\end{abstract}
\section{Introduction}
Meteorological downscaling is a core technology that correlates the low-resolution output of Global Climate Models (GCMs) with high-precision regional meteorological field forecasts. It provides crucial support for climate change assessment, extreme weather early warning, and ecological environment governance. The frequency of global extreme meteorological events is increasing, and the demand for high-resolution and physically consistent meteorological data is becoming increasingly urgent. The accuracy of downscaling results directly determines the reliability of disaster risk assessment and emergency decision-making.
Traditional dynamic downscaling employs Regional Climate Models (RCMs) that share similar physical processes with GCMs,  
but have higher resolution and smaller coverage. It can effectively simulate extreme values, but incurs large computational costs and requires professional knowledge for result interpretation. Statistical downscaling is another downscaling approach. With the development of deep learning, leveraging data-driven capabilities for meteorological downscaling reduces substantial computational burdens and minimizes reliance on manual prior knowledge. 
Existing downscaling methods operating at fixed scales lead to error propagation when acquiring information at arbitrary scales \cite{liu2024arbitrary}. Meanwhile, operator learning naturally possesses the ability to make predictions across arbitrary scales as it focuses on learning mappings between function spaces, and it has been successfully applied to tasks related to partial differential equations in fields like fluid mechanics and aerodynamics \cite{xiao2024fourier,PEYVAN2026114432}.

\begin{figure}[t]
    \centering
    \includegraphics[width=0.62\linewidth, angle=-90]{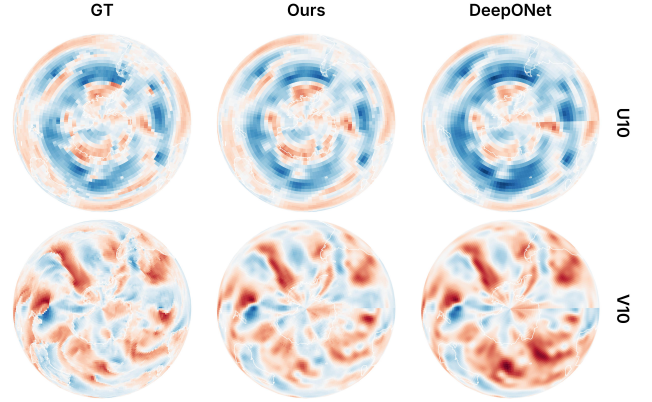} 
    \caption{Visualization of climate downscaling results at $2\times$ (training) and $4\times$ (OOD) factors. MSPNO generalizes across variables and scales while preserving spherical continuity and mitigating polar discontinuities.}
    \label{fig:climate_placeholder}
\end{figure}

\begin{table}[t]
    \centering
    \footnotesize
    \begin{tabular}{l c c c}
        \toprule
        \textbf{Method} & 
        \textbf{Multi-scale} &
        \textbf{Sphere} &
        \textbf{Physics} \\
        \midrule
        \multicolumn{4}{l}{\textit{Class 1: Dot-product operators}} \\
        DeepONet, LOCA    & \ding{55} & \ding{55} & \ding{55} \\
        \cmidrule(lr){1-4} 
        \multicolumn{4}{l}{\textit{Class 2: Fourier-based operator}} \\
        FNO        & \ding{55} & \ding{55} & \ding{55} \\
        SFNO       & \ding{55} & \ding{51} & \ding{55} \\
        \cmidrule(lr){1-4}
        \multicolumn{4}{l}{\textit{Class 3: Transformer-based operator}} \\
        OFormer, LNO, IPOT    & \ding{55} & \ding{55} & \ding{55} \\
        GNOT       & \ding{51} & \ding{55} & \ding{55} \\
        \cmidrule(lr){1-4}
        \textbf{MSPNO (Ours)} & \ding{51} & \ding{51} & \ding{51} \\
        \bottomrule
    \end{tabular}
    \caption{Conceptual comparison of neural operator methods.}
    \label{tab:comparison}
\end{table}

Despite this progress, operator learning for meteorological downscaling faces three fundamental challenges, as also reflected in Table~\ref{tab:comparison}.
First, meteorological systems exhibit intensive cross-scale coupling characteristics \cite{frank2024characterizing}. Global-scale atmospheric circulations, mesoscale weather patterns, and localized convective processes do not exist in isolation; rather, they are fundamentally intertwined through complex non-linear interactions and reciprocal dynamical feedback loops. This intricate cross-scale correlation significantly increases the difficulty of operator learning in capturing the comprehensive spectrum of atmospheric dynamics across disparate spatial frequencies.
Second, meteorological data are based on non-uniform and distorted spherical grid structures (e.g., grid cells in high-latitude regions show significant compression). As illustrated in Fig.~\ref{fig:climate_placeholder}, standard operators like DeepONet~\cite{lu2021learning} often suffer from severe polar discontinuities, necessitating geometry-consistent manifold modeling. The core challenge lies in the need for the model to fully adapt to and utilize the inherent properties of spherical manifolds, effectively capture valid information around query points, and thereby address the representational biases caused by grid non-uniformity and distortion.
Third, the integration of physical consistency remains elusive. Purely data-driven operators often produce predictions that violate fundamental conservation laws, necessitating the embedding of differentiable physical priors to ensure dynamical realism.

Early operator learning methods, such as DeepONet~\cite{lu2021learning} and its decoupled variants like LOCA~\cite{kissas2022learning}, provide approximation guarantees under specific assumptions but often suffer from high training sensitivity and performance saturation due to their inherent linear operator structures. 
Spectral-type operators, including FNO~\cite{DBLP:conf/iclr/LiKALBSA21} and SFNO~\cite{bonev2023spherical}, accelerate inference via Fourier-domain convolutions; however, they inherently lack arbitrary-scale prediction capabilities and require additional post-processing for scale adaptation~\cite{li2023fourier}. 
Subsequent Transformer-based operators, such as GNOT~\cite{hao2023gnot}, LNO~\cite{wang2024LNO}, and IPOT~\cite{lee2024inducing}, leverage strong inductive biases but typically necessitate massive datasets for scale-aware learning, while lacking tailored priors for spherical geometry or atmospheric multi-scale features. 
Although recent physics-integrated models like NeuralGCM~\cite{kochkov2024neural} and ClimODE~\cite{verma2024climode} target global weather forecasting, and PIANO~\cite{chin2025piano} focuses on precipitation, none are explicitly optimized for the dual requirements of meteorological downscaling: accommodating spherical topology while enforcing dynamical consistency across coupled spatial scales.

In this paper, we propose \textbf{MSPNO}, a \textbf{M}ulti-scale \textbf{S}pherical \textbf{P}hysics-informed \textbf{N}eural \textbf{O}perator for meteorological downscaling. To capture hierarchical atmospheric dynamics, MSPNO employs a multi-scale encoder based on spherical Laplacian decomposition. A geometry-consistent query interface is established on the $\mathbb{S}^2$ manifold through hybrid geodesic coordinate encoding. Recognizing that meteorological downscaling is inherently a localized reconstruction task driven by neighborhood contexts, we formulate a localized spherical integral operator that exploits geodesic sparsity. This design adheres to the task-specific requirement of modeling local spatial dependencies while maintaining efficient, grid-agnostic inference. Finally, differentiable physics-informed constraints are embedded into the learning objective to ensure dynamical consistency and structural fidelity across the predicted atmospheric fields.

\section{Related Work}
\subsection{Meteorological Downscaling}

Downscaling methods have long been applied in climate science. 
Advances in deep learning have spurred the adoption of diverse models for meteorological downscaling. FSRCNN \cite{dong2016accelerating} and RCAN \cite{zhang2018image}, adapted from super-resolution tasks, improve reconstruction accuracy but fail to adapt to meteorological variables’ unique properties. Later climate-tailored models like DeepSD \cite{vandal2017deepsd} and YNet \cite{liu2020climate} enhance feature fusion for climate variables, and GINE \cite{park2022downscaling} incorporates terrain information—yet none enable arbitrary-scale prediction. Even recent architectures tailored to meteorological scenes’ spherical nature, such as SFNO \cite{bonev2023spherical} and SCNN \cite{cohen2018spherical}, remain constrained by a key bottleneck: lacking both arbitrary-scale prediction and multi-scale adaptability for meteorological tasks. We therefore address this problem via a neural operator functionally adapted to meteorological characteristics.

\subsection{Neural Operators}
Neural operators model function-to-function mappings and are naturally suited for meteorological downscaling. Representative methods include DeepONet~\cite{lu2021learning} with branch-trunk decomposition while LOCA~\cite{kissas2022learning} enhances branches via explicit kernels yet remains a linear operator with limited expressive power. Fourier-type operators such as FNO~\cite{DBLP:conf/iclr/LiKALBSA21} and SFNO~\cite{bonev2023spherical} require post-processing to achieve arbitrary-scale prediction~\cite{li2023fourier}. Subsequent works leverage Transformer architectures for operator learning: OFormer~\cite{li2022transformer} directly inherits the Transformer Encoder-Decoder structure LNO~\cite{wang2024LNO} enhances generalization by sharing input and query coordinate encodings and IPOT~\cite{lee2024inducing} adopts a lightweight attention architecture for acceleration. Meanwhile some studies combine PINNs with neural operators to improve task-specific loss functions. However, adapting these operators to global meteorological fields remains non-trivial. Currently a downscaling-oriented operator is still missing that (i) accommodates the inherent multi-scale features and spherical topology of meteorological data and (ii) is designed to be compatible with physical consistency constraints during training.

\section{Method}
\begin{figure*}[t]
    \centering
    \includegraphics[width=0.95\linewidth]{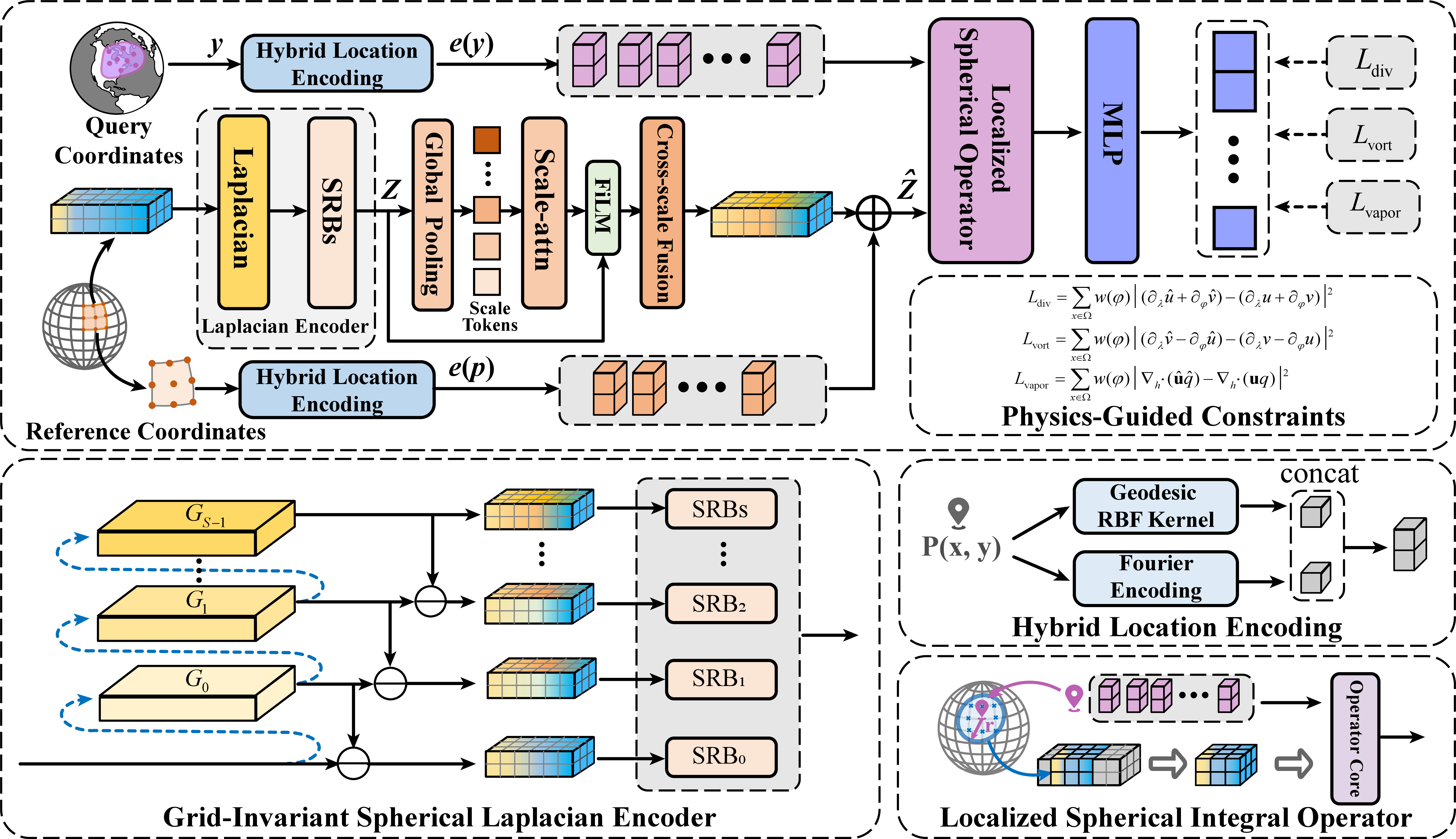} 
    \caption{Overview. The \textbf{Spherical Laplacian Encoder} hierarchically decomposes multi-scale fluctuations on the spherical manifold. The \textbf{Hybrid Location Encoding} combines geodesic RBF kernels and Fourier features for continuous query evaluation. The \textbf{Localized Spherical Integral Operator} enables grid-agnostic evaluation through geometry-aware attention, while the \textbf{Cross-scale Fusion} module calibrates multi-frequency representations. Finally, differentiable \textbf{Physics-Guided Constraints} enforce dynamical consistency and physical fidelity.}
    \label{fig:model}
\end{figure*}

\subsection{Problem Formulation}
Let \( \mathbb{S}^2 \subset \mathbb{R}^3 \) denote the 2-dimensional sphere. 
Let \( H, W \) be positive integers denoting the height and width of the discrete sampling grid on \( \mathbb{S}^2 \). 
Given \( N \) pairs of input-output data \( \{u^\ell, v^\ell(y)\}_{\ell=1}^N \) generated by an unknown ground truth operator 
\( \mathcal{G}: \mathbb{R}^{d_u \times H \times W} \to C(\mathbb{S}^2, \mathbb{R}^{d_v}) \), 
where \( u^\ell \in \mathbb{R}^{d_u \times H \times W} \) represents \( d_u \)-dimensional input variables sampled on the 
\( H \times W \) grid of \( \mathbb{S}^2 \), 
\( v^\ell \in C(\mathbb{S}^2, \mathbb{R}^{d_v}) \) denotes \( d_v \)-dimensional output functions defined on \( \mathbb{S}^2 \), 
and \( y \in \mathbb{S}^2 \) is a query location on \( \mathbb{S}^2 \). 
Our goal is to learn an operator 
\( \mathcal{F}: \mathbb{R}^{d_u \times H \times W} \to C(\mathbb{S}^2, \mathbb{R}^{d_v}) \) 
such that for \( \ell = 1, \dots, N \),
\[
\mathcal{F}(u^\ell) = v^\ell.
\]

\subsection{Grid-Invariant Spherical Laplacian Encoder}
\label{sec:spherical_encoder}

Let $u \in \mathbb{R}^{d_u \times H \times W}$ denote an input
spherical field sampled on a fixed grid over $\mathbb{S}^2$.
Inspired by \cite{zhuang2025ladcast}, we construct a sequence of scale-dependent spherical filters $\{G_s\}_{s=0}^{S-1}$ . Each $G_s$ is implemented as a spherical convolution on $\mathbb{S}^2$, specifically designed to account for the spherical metric and mitigate coordinate-induced distortions, particularly the convergence of meridians in polar regions.
The receptive field of $G_s$ increases monotonically with the scale index $s$, inducing progressively stronger low-pass filtering on the sphere without any form of spatial subsampling.

Using these filters, we define a Spherical Laplacian Decomposition:
\begin{align}
u_0 &= u, \\
\mathbf{L}_s &= u_s - G_s(u_s), \quad s = 0, \dots, S-2, \\
u_{s+1} &= G_s(u_s),
\end{align}
where $\mathbf{L}_s \in \mathbb{R}^{d_u \times H \times W}$ represents the
scale-specific residual at scale $s$, and the final component
$\mathbf{L}_{S-1} = u_{S-1}$ captures the global low-frequency
structure.
This manifold-consistent decomposition decouples localized fluctuations from global atmospheric modes while mitigating artifacts induced by meridian convergence. By avoiding spatial downsampling, it preserves exact point-wise correspondence across all scales, aligning fine-grained details and large-scale structures within a unified spherical representation.

To model dependencies across characteristic spatial scales, each hierarchical band $\mathbf{L}_s$ (initially in $\mathbb{R}^{d_u \times H \times W}$) is processed by a Spherical Residual Block (SRB), 
which is a residual block with spherical kernels and it transforms the multi-scale residuals into a deep latent feature map $\mathbf{z}_s \in \mathbb{R}^{d_c \times H \times W}$, where $d_c$ denotes the hidden convolutional dimension. We then condense these scale-specific representations into compact scale tokens $\mathbf{t}_s \in \mathbb{R}^{d_c}$ using latitude-weighted spherical global average pooling:
\begin{equation}
\mathbf{t}_s = \frac{1}{Z} \sum_{h=1}^{H} \sum_{w=1}^{W} \cos(\varphi_h)\, \mathbf{z}_s(h,w),
\end{equation}
where $\mathbf{z}_s(h,w) \in \mathbb{R}^d_c$ denotes the feature vector at location $(h,w)$. The term $\cos(\varphi_h)$ accounts for the spherical area element associated with each grid cell, and the normalization constant is defined as $Z = \sum_{h,w} \cos(\varphi_h)$.
The resulting scale-level sequence $\mathbf{T} = \{\mathbf{t}_s\}_{s=0}^{S-1}$ is processed by a Scale-Aware Attention to capture global interactions across frequencies, yielding contextualized tokens $\mathbf{t}_s^c$.

To dynamically calibrate the importance of different spatial frequencies, we employ Feature-wise Linear Modulation (FiLM) \cite{perez2018film}. Specifically, for each scale $s$, the contextualized token $\mathbf{t}_s^c$ is projected through learnable linear layers to generate scale-specific affine parameters $\boldsymbol{\gamma}_s, \boldsymbol{\beta}_s \in \mathbb{R}^{d_c}$. These vectors are then applied to modulate the corresponding feature map $\mathbf{z}_s$ across the channel dimension:
\begin{equation}
\tilde{\mathbf{z}}_s(h,w) = \boldsymbol{\gamma}_s \odot \mathbf{z}_s(h,w) + \boldsymbol{\beta}_s,
\end{equation}
where $\odot$ denotes element-wise multiplication. 

We further aggregate the modulated features across the scale dimension via a point-wise cross-scale convolution module, yielding a unified representation $\mathbf{z} \in \mathbb{R}^{d_c \times H \times W}$. 
Finally, a pixel-wise projection maps this to the latent sequence $\mathbf{Z} \in \mathbb{R}^{L \times D}$, where $L = H \times W$.

\subsection{Hybrid Location Encoding}
To bridge discrete latent fields with the continuous $\mathbb{S}^2$ manifold, we define a geometry-aware query representation that couples spectral coordinate embeddings with geodesic radial basis function (RBF) interpolation. This formulation enables grid-agnostic operator evaluation, allowing the model to query atmospheric states at arbitrary spherical locations.

Each query location $y \in \mathbb{S}^2$ is parameterized by its spherical coordinates $(\varphi, \lambda)$ and mapped into a high-dimensional spectral space via Fourier features:
\begin{equation}
    \phi(y)
    =
    \left[
        \cos\!\left([\varphi, \lambda]^\top \mathbf{W}_\phi\right)
        \;;\;
        \sin\!\left([\varphi, \lambda]^\top \mathbf{W}_\phi\right)
    \right],
\end{equation}
where $\mathbf{W}_\phi$ is a learnable projection that enables the model to capture multi-frequency spatial priors. 

To anchor the continuous query point to the discrete latent field, we interpolate the reference latents $\{\mathbf{z}_i\}$ located at $\{p_i\}$ using a normalized geodesic RBF kernel:
\begin{equation}
    \tilde{\mathbf{z}}(y)
    =
    \sum_i
    \frac{\exp\!\left(-\epsilon d_{\mathbb{S}^2}^2(y,p_i)\right)}
         {\sum_j \exp\!\left(-\epsilon d_{\mathbb{S}^2}^2(y,p_j)\right)}
    \mathbf{z}_i,
\end{equation}
where $d_{\mathbb{S}^2}(\cdot,\cdot)$ is the spherical geodesic distance and $\epsilon$ is a learnable bandwidth parameter.

The final query embedding is given by
\begin{equation}
    \mathbf{e}(y)
    =
    \mathrm{MLP}\!\left([\phi(y)\,;\,\tilde{\mathbf{z}}(y)]\right),
\end{equation}
and is applied identically to continuous query coordinates and reference grid
points, yielding a unified, geometry-consistent query interface for subsequent
operator evaluation.

\subsection{Localized Spherical Integral Operator}


We first define a localized spherical integral operator $\mathcal{F}$ that maps the latent field to a query location via a localized integral form inspired by neighborhood attention on the sphere \cite{bonev2026attention}:
\begin{equation}
    \mathcal{F}[z](y) = \int_{\mathbb{S}^2} \kappa(y, x) v(x) \, d\mu(x),
\end{equation}
where $d\mu(x)$ is the spherical measure. The kernel $\kappa(y, x)$ represents a data-dependent similarity between the query location $y$ and a source location $x$. It is defined using a localized softmax-normalization:
\begin{equation}
    \kappa(y, x) = \frac{I_{D(y)}(x) \exp\left( Q(y) \cdot K(x) / \sqrt{d} \right)}{\int_{\mathbb{S}^2} I_{D(y)}(x') \exp\left( Q(y) \cdot K(x') / \sqrt{d} \right) d\mu(x')},
\end{equation}
where $Q(y)$ is the query encoding derived from the geometry of $y$, while $K(x)$ is the key encoding of the source point $x$. The indicator function $I_{D(y)}(x)$ restricts the kernel support to a local spherical disk $D(y)$ of radius $\theta$.

In practice, the latent field is available only as a finite set of embeddings $\mathbf{Z} = \{\mathbf{z}_i\}_{i=1}^L$ at discrete reference locations $\{p_i\}_{i=1}^L$. We implement the operator by approximating the continuous integral with a discrete summation over the sampled reference points $p_i \in D(y)$:
\begin{equation}
    \mathcal{F}[\mathbf{z}](y) \approx \sum_{i \in \mathcal{N}(y)} \alpha_i(y) \mathbf{W}_v \hat{\mathbf{z}}_i,
\end{equation}
where $\mathcal{N}(y) = \{i \mid d_{\mathbb{S}^2}(y, p_i) \le \theta\}$ is the set of indices of reference points within the query's local neighborhood. Here, $\mathbf{W}_v \hat{\mathbf{z}}_i$ serves as the discrete realization of the value field $v(p_i)$ at the reference points. The discrete attention weights $\alpha_i(y)$ are obtained by sampling and re-normalizing the continuous kernel $\kappa(y, x)$ at the reference locations $p_i$:
\begin{equation}
    \alpha_i(y) = \frac{\exp\left( Q(y) \cdot K(p_i) / \sqrt{d} \right)}{\sum_{j \in \mathcal{N}(y)} \exp\left( Q(y) \cdot K(p_j) / \sqrt{d} \right)}.
\end{equation}
This formulation ensures the operator preserves the underlying continuous dynamics while remaining efficiently computable. 

Finally, the output of the operator is passed through a channel-wise linear readout to produce the atmospheric field predictions $C(\mathbb{S}^2,\mathbb{R}^{d_v})$ at the specified query locations.

\subsection{Physics-Guided Learning Objective}

To enhance physical consistency beyond purely data-driven training, we
incorporate physics-guided constraints directly into the learning objective.

\paragraph{Latitude-weighted regression loss.}
Given predictions $\hat{y}$ and ground truth $y$ over $V$ variables, we employ
a latitude-weighted mean squared error:
\begin{equation}
\mathcal{L}_{\mathrm{MSE}}
=
\frac{1}{|\Omega|}
\sum_{v=1}^{V}
\sum_{(\varphi,\lambda)\in\Omega}
w(\varphi)
\bigl(
\hat{y}_{v}(\varphi,\lambda)
-
y_{v}(\varphi,\lambda)
\bigr)^2 .
\end{equation}
\begin{table*}[htb]
\centering
\renewcommand{\arraystretch}{1.15}

\begin{tabular}{ccrrrrrrr}
\toprule
\textbf{Metric} & \textbf{Variable} & \textbf{Ours} & \textbf{OFormer} & \textbf{GNOT} & \textbf{LNO} & \textbf{DeepONet} & \textbf{IPOT} & \textbf{LOCA} \\
\midrule

\multirow{8}{*}{\makecell{MSE\\($\downarrow$)}} 
 & T2m  & \textbf{0.0045} & \underline{0.0066} & 0.0117 & 0.0117 & 0.0127 & 0.0105 & 0.0102 \\
 & U10  & \textbf{0.1293} & \underline{0.1397} & 0.1478 & 0.1478 & 0.2182 & 0.1471 & 0.1477 \\
 & V10  & \textbf{0.1637} & \underline{0.1763} & 0.1870 & 0.1868 & 0.3871 & 0.1847 & 0.1865 \\
 & Z500 & \textbf{0.0009} & \underline{0.0009} & 0.0010 & 0.0010 & 0.0034 & \underline{0.0009} & 0.0010 \\
 & T850 & \textbf{0.0064} & \underline{0.0074} & 0.0076 & 0.0076 & 0.0099 & 0.0075 & 0.0075 \\
 & R500 & 0.3526 & \textbf{0.3505} & \underline{0.3511} & \underline{0.3511} & 0.5380 & 0.3524 & 0.3512 \\
 & S850 & \textbf{0.1250} & \underline{0.1277} & 0.1324 & 0.1323 & 0.1421 & 0.1319 & 0.1315 \\
 & \textbf{Average} & \textbf{0.1117} & \underline{0.1156} & 0.1198 & 0.1197 & 0.1873 & 0.1193 & 0.1194 \\

\midrule

\multirow{8}{*}{\makecell{RMSE\\($\downarrow$)}} 
 & T2m  & \textbf{1.4322} & \underline{1.7125} & 2.3027 & 2.3021 & 2.3951 & 2.1806 & 2.1457 \\
 & U10  & \textbf{1.9831} & \underline{2.0614} & 2.1208 & 2.1209 & 2.5368 & 2.1159 & 2.1202 \\
 & V10  & \textbf{1.9225} & \underline{1.9944} & 2.0547 & 2.0537 & 2.8609 & 2.0423 & 2.0523 \\
 & Z500 & \textbf{102.8555} & \underline{106.7649} & 109.7369 & 109.7392 & 195.1579 & 109.6546 & 108.9631 \\
 & T850 & \textbf{1.2558} & \underline{1.3503} & 1.3657 & 1.3658 & 1.5556 & 1.3594 & 1.3537 \\
 & R500 & 19.8951 & \textbf{19.8380} & \underline{19.8525} & 19.8546 & 24.2521 & 19.8915 & 19.8576 \\
 & S850 & \textbf{0.00148} & \underline{0.00150} & \underline{0.00150} & \underline{0.00150} & 0.00160 & \underline{0.00150} & \underline{0.00150} \\
 & \textbf{Average} & \textbf{18.4779} & \underline{19.1033} & 19.6335 & 19.6340 & 32.6814 & 19.6052 & 19.4991 \\

\midrule

\multirow{8}{*}{\makecell{ACC\\($\uparrow$)}} 
 & T2m  & \textbf{0.9622} & \underline{0.9462} & 0.9078 & 0.9156 & 0.9214 & 0.9365 & 0.9388 \\
 & U10  & \textbf{0.8819} & \underline{0.8720} & 0.8653 & 0.8659 & 0.8142 & 0.8658 & 0.8664 \\
 & V10  & \textbf{0.8907} & \underline{0.8820} & 0.8754 & 0.8762 & 0.7774 & 0.8766 & 0.8772 \\
 & Z500 & \textbf{0.9948} & 0.9943 & 0.9940 & 0.9942 & 0.9819 & \underline{0.9944} & 0.9947 \\
 & T850 & \underline{0.9683} & 0.9633 & 0.9626 & 0.9644 & 0.9583 & 0.9685 & \textbf{0.9694} \\
 & R500 & 0.7293 & \textbf{0.7305} & 0.7301 & \underline{0.7303} & 0.6375 & 0.7298 & \underline{0.7303} \\
 & S850 & \underline{0.7897} & 0.7848 & 0.7784 & 0.7811 & 0.7735 & 0.7831 & \textbf{0.7934} \\
 & \textbf{Average} & \textbf{0.8881} & \underline{0.8819} & 0.8734 & 0.8754 & 0.8377 & 0.8792 & 0.8815 \\

\bottomrule
\end{tabular}

\caption{Quantitative comparison of different operator learning models in terms of MSE, RMSE, and ACC across multiple atmospheric variables. Best results are highlighted in \textbf{bold}, with second-best underlined.}
\label{tab::main}
\end{table*}

\paragraph{Physics-guided constraints.}
To encourage consistency with large-scale atmospheric dynamics, we introduce
a set of differentiable physics-guided constraints defined on a latitude--longitude
parameterization of spherical domain.
Spatial derivatives with respect to longitude $\lambda$ and latitude $\varphi$
are computed using spectral differentiation on the discrete grid.

Let $\hat{u}=(\hat{u},\hat{v})$ denote the predicted horizontal wind,
$u=(u,v)$ the ground truth wind, and $\hat{q},q$ the predicted and
true specific humidity.
We define three physically motivated residuals:
\begin{align}
\mathcal{L}_{\mathrm{div}} &=
\sum_{x\in\Omega}
w(\varphi)
\bigl|
(\partial_\lambda \hat{u} + \partial_\varphi \hat{v})
-
(\partial_\lambda u + \partial_\varphi v)
\bigr|^2 , \\
\mathcal{L}_{\mathrm{vort}} &=
\sum_{x\in\Omega}
w(\varphi)
\bigl|
(\partial_\lambda \hat{v} - \partial_\varphi \hat{u})
-
(\partial_\lambda v - \partial_\varphi u)
\bigr|^2 , \\
\mathcal{L}_{\mathrm{vapor}} &=
\sum_{x\in\Omega}
w(\varphi)
\bigl|
\nabla_h\!\cdot(\hat{u}\hat{q})
-
\nabla_h\!\cdot(uq)
\bigr|^2 ,
\end{align}
where $\nabla_h\!\cdot(uq)=\partial_\lambda(uq)+\partial_\varphi(vq)$
denotes a discrete approximation of the horizontal moisture flux divergence.
The raw physics loss is a weighted combination of the above terms:
\begin{equation}
\mathcal{L}_{\mathrm{phys}}^{\mathrm{raw}}
=
\lambda_{\mathrm{div}}\mathcal{L}_{\mathrm{div}}
+
\lambda_{\mathrm{vort}}\mathcal{L}_{\mathrm{vort}}
+
\lambda_{\mathrm{vapor}}\mathcal{L}_{\mathrm{vapor}} .
\end{equation}

\paragraph{Normalization and total objective.}
To balance the magnitude of the physics constraints with the regression loss,
we rescale $\mathcal{L}_{\mathrm{phys}}^{\mathrm{raw}}$ using an exponential
moving average, yielding $\mathcal{L}_{\mathrm{phys}}^{(t)}$.
A dynamic weight $\lambda_{\mathrm{phys}}^{(t)}$ is further applied to avoid
over-regularization during early training.
The final objective at training step $t$ is
\begin{equation}
\mathcal{L}_{\mathrm{total}}^{(t)}
=
\mathcal{L}_{\mathrm{MSE}}^{(t)}
+
\lambda_{\mathrm{phys}}^{(t)}\,
\mathcal{L}_{\mathrm{phys}}^{(t)} .
\end{equation}

\section{Numerical Experiments}



This section evaluates the proposed physics-constrained multi-scale spherical operator through statistical downscaling benchmarks across multiple variables, zero-shot generalization tests across varied resolution ratios, and comprehensive ablation studies. By analyzing scale-wise representations and error distributions, particularly in geographically sensitive areas, we aim to address the following key questions\footnote{\url{https://github.com/YanisYe/MSPNO}}:

\begin{itemize}
    \item \textbf{Q1:} How do explicit spherical multi-scale operators improve prediction accuracy for high-resolution forecasting compared to existing methods?

    \item \textbf{Q2:} Can the proposed model consistently maintain robustness and accuracy when generalizing across different meteorological downscaling factors?

    \item \textbf{Q3:}  Does physics-informed constraints and spherical locality contribute to stabilizing the learning process and enhancing physical fidelity?
\end{itemize}


\subsection{Experiment setup and evaluation details}

\paragraph{Datasets.} 
We utilize the preprocessed ERA5 reanalysis dataset from WeatherBench \cite{rasp2020weatherbench}. 
For training, we employ data spanning the years (2014--2015), subsampled at a 6-hour temporal interval. We consider an input dimensionality of $d_u = 54$ and an output dimensionality of $d_s = 7$. The target variables encompass surface fields, including 2m temperature (\texttt{T2m}), 10m wind components (\texttt{U10}, \texttt{V10}), and upper-air fields (\texttt{Z500}, \texttt{T850}, \texttt{R500}, \texttt{S850}), More details in Appendix~A.

\paragraph{Metrics.} 
We evaluate downscaling performance using latitude-weighted Mean Squared Error (MSE), Root Mean Squared Error (RMSE), and the Anomaly Correlation Coefficient (ACC). Refer to Appendix~B for formal definitions.

\paragraph{Baselines.} 
We compare our proposed model with several representative state-of-the-art method including DeepONet~\cite{lu2021learning}, GNOT~\cite{hao2023gnot}, LNO~\cite{wang2024LNO}, OFormer~\cite{li2022transformer}, IPOT~\cite{lee2024inducing} and LOCA~\cite{kissas2022learning}. More details in Appendix~C.

\subsection{Main Results}
Quantitative results are summarized in Table~\ref{tab::main} using MSE, RMSE, and ACC.
Our method consistently outperforms existing operator-learning baselines across all evaluation metrics, achieving the lowest average MSE (\textbf{0.1117}) and RMSE (\textbf{18.4779}) alongside the highest ACC (\textbf{0.8881}). 
These results indicate that our model delivers more accurate and stable
downscaling across diverse atmospheric variables under a unified evaluation setting.
The improvements are evident for dynamically complex fields such as near-surface winds and geopotential height, suggesting that the proposed combination of explicit modeling of spherical geometry, coupled with multi-scale Laplacian decomposition, and embedded physical constraints is especially effective for capturing cross-scale atmospheric structures in realistic downscaling scenarios.
\begin{table}[t]
\centering
\renewcommand{\arraystretch}{1.1}
\begin{tabular}{lccc}
\toprule
\textbf{Scales} & \textbf{MSE ($\downarrow$)} & \textbf{RMSE ($\downarrow$)} & \textbf{ACC ($\uparrow$)} \\
\midrule
1 (non-hierarchical) & 0.1125 & 18.5787 & 0.8876 \\
4 (Ours) & \textbf{0.1117} & \textbf{18.4779} & 0.8881 \\
8 & \textbf{0.1117} & 18.5175 & \textbf{0.8882} \\
\bottomrule
\end{tabular}
\caption{Hyper-parameter study on the number of spatial scales in the spherical encoder. Results show that a 4-scale decomposition provides the optimal balance between accuracy and complexity.}
\label{tab:spatial_scales}
\end{table}

\begin{table*}[htb]
\centering
\small
\renewcommand{\arraystretch}{1.15}
\begin{tabular}{llrrrrrrrr}
\toprule
\textbf{Metric} & \textbf{Model}
& T2m & U10 & V10 & Z500 & T850 & R500 & S850 & \textbf{Average} \\
\midrule

\multirow{4}{*}{MSE ($\downarrow$)}
& Ours
& 0.0045 & 0.1293 & 0.1637 & 0.0009 & 0.0064 & 0.3526 & 0.1250 & \textbf{0.1117} \\
& w/ Encoder (repl. Sphere)
& 0.0045 & 0.1298 & 0.1644 & 0.0009 & 0.0064 & 0.3540 & 0.1250 & 0.1121 \\
& w/ Attn. Kernel (repl. Sphere)
& 0.0050 & 0.1323 & 0.1679 & 0.0009 & 0.0065 & 0.3511 & 0.1252 & 0.1127 \\
& w/o Physics-Guided Objective
& 0.0046 & 0.1295 & 0.1642 & 0.0009 & 0.0064 & 0.3531 & 0.1249 & 0.1119 \\

\midrule

\multirow{4}{*}{RMSE ($\downarrow$)}
& Ours
& 1.4322 & 1.9831 & 1.9225 & 102.8555 & 1.2558 & 19.8951 & 0.0015 & \textbf{18.4779} \\
& w/ Encoder (repl. Sphere)
& 1.4308 & 1.9871 & 1.9261 & 103.3402 & 1.2575 & 19.9353 & 0.0015 & 18.5541 \\
& w/ Attn. Kernel (repl. Sphere)
& 1.5079 & 2.0060 & 1.9465 & 104.9548 & 1.2661 & 19.8532 & 0.0015 & 18.7909 \\
& w/o Physics-Guided Objective
& 1.4390 & 1.9849 & 1.9252 & 103.1875 & 1.2572 & 19.9113 & 0.0015 & 18.5288 \\

\midrule

\multirow{4}{*}{ACC ($\uparrow$)}
& Ours
& 0.9622 & 0.8819 & 0.8907 & 0.9948 & 0.9683 & 0.7293 & 0.7897 & \textbf{0.8881} \\
& w/ Encoder (repl. Sphere)
& 0.9624 & 0.8815 & 0.8903 & 0.9947 & 0.9682 & 0.7284 & 0.7897 & 0.8879 \\
& w/ Attn. Kernel (repl. Sphere)
& 0.9584 & 0.8791 & 0.8880 & 0.9946 & 0.9678 & 0.7304 & 0.7894 & 0.8868 \\
& w/o Physics-Guided Objective
& 0.9621 & 0.8817 & 0.8905 & 0.9947 & 0.9682 & 0.7290 & 0.7897 & 0.8880 \\

\bottomrule
\end{tabular}
\caption{Ablation study verifying spherical components and physics-guided objectives.
}
\label{tab:ablation_kernel_physics}
\end{table*}

\subsection{Zero-Shot Generalization}
\begin{figure}[t]
\centering
\includegraphics[width=0.98\linewidth]{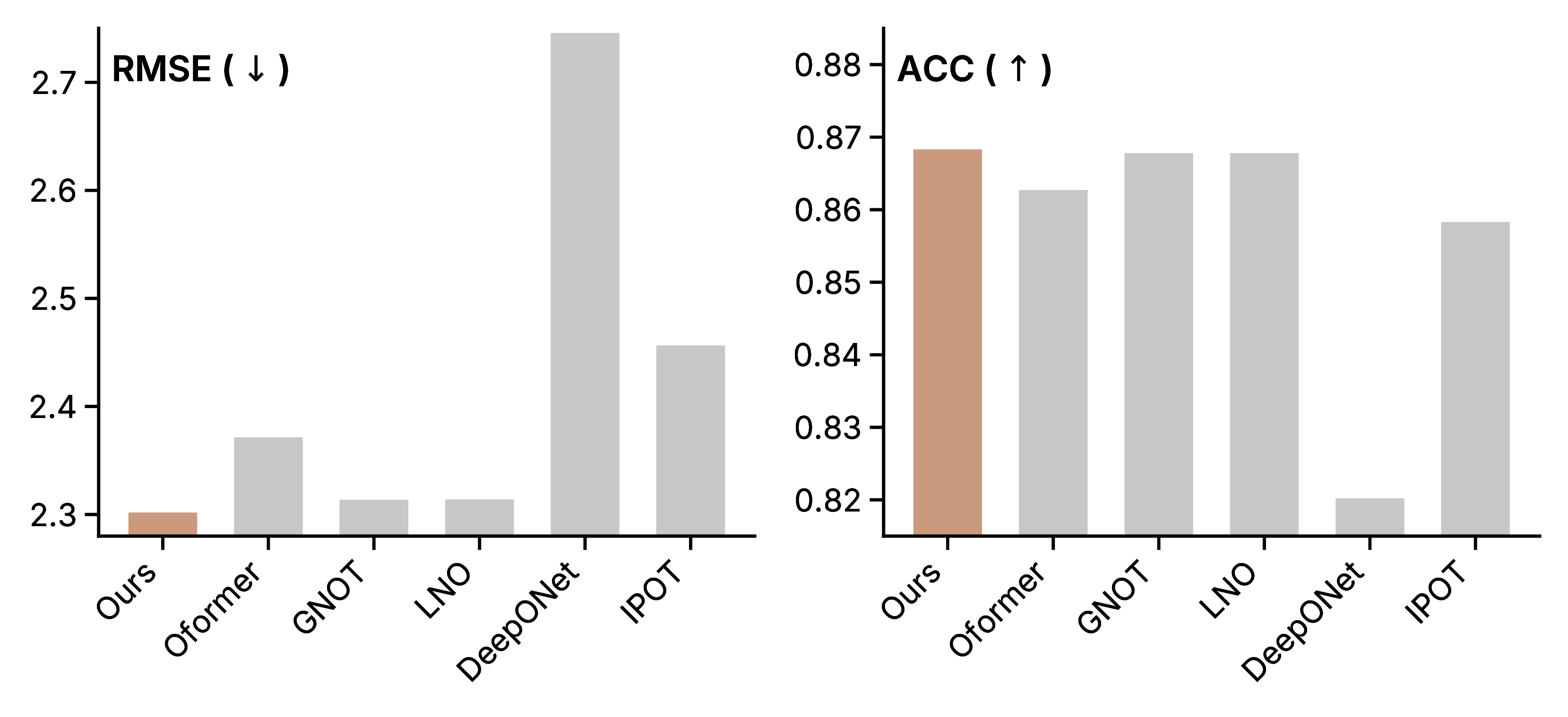} 
\caption{Zero-shot performance comparison across unseen query locations. }
\label{fig:zeroshot_bar}
\end{figure}

A key advantage of neural operators is their ability to perform grid-agnostic inference. We evaluate this by applying models trained on a $2\times$ downscaling task ($5.625^\circ \to 2.8125^\circ$) directly to a $4\times$ task ($5.625^\circ \to 1.40625^\circ$) without further fine-tuning. This zero-shot setting assesses the model's capacity to extrapolate atmospheric dynamics to out-of-distribution (OOD) resolutions.
\begin{table}[t]
\centering
\small
\renewcommand{\arraystretch}{1.1}
\begin{tabular}{lcccc}
\toprule
\textbf{Method} & \textbf{Div.} $\downarrow$ & \textbf{Vort.} $\downarrow$ & \textbf{Vapor} $\downarrow$ & \textbf{Total} $\downarrow$ \\
\midrule
\textbf{Ours} & \textbf{189.0197} & \textbf{78.5133} & \textbf{160.9526} & \textbf{428.4856} \\
OFormer & 202.2805 & 84.1339 & 171.9428 & 458.3572 \\
GNOT & 213.3368 & 88.0060 & 173.7128 & 475.0552 \\
LNO & 213.2525 & 87.9770 & 173.8381 & 475.0676 \\
DeepONet & 218.6825 & 89.9197 & 183.1225 & 491.7247 \\
IPOT & 211.7478 & 87.3884 & 173.8754 & 463.0116 \\
LOCA & 213.1982 & 87.9701 & 174.0555 & 475.2237 \\
\bottomrule
\end{tabular}
\caption{Physical consistency evaluation on $2\times$ downscaling. We report mean absolute residuals ($\times 10^{-5}$) for divergence, vorticity, and vapor transport.}
\label{tab:physical_consistency}
\end{table}

As shown in Fig.~\ref{fig:zeroshot_bar}, our method achieves the best
average RMSE and ACC under this unseen downscaling factor.
In contrast, operator-learning baselines such as OFormer~\cite{li2022transformer}, GNOT~\cite{hao2023gnot}, and LNO~\cite{wang2024LNO} exhibit
larger error increases and reduced accuracy when extrapolating to finer
resolutions.
These results indicate that our synergistic architectural design fosters stable, scale-consistent representations, enabling reliable zero-shot extrapolation to finer, out-of-distribution resolutions.

\subsection{Effectiveness and Interpretation of Multi-scale Representations}

\begin{figure}[t]
\centering
\includegraphics[width=0.98\linewidth]{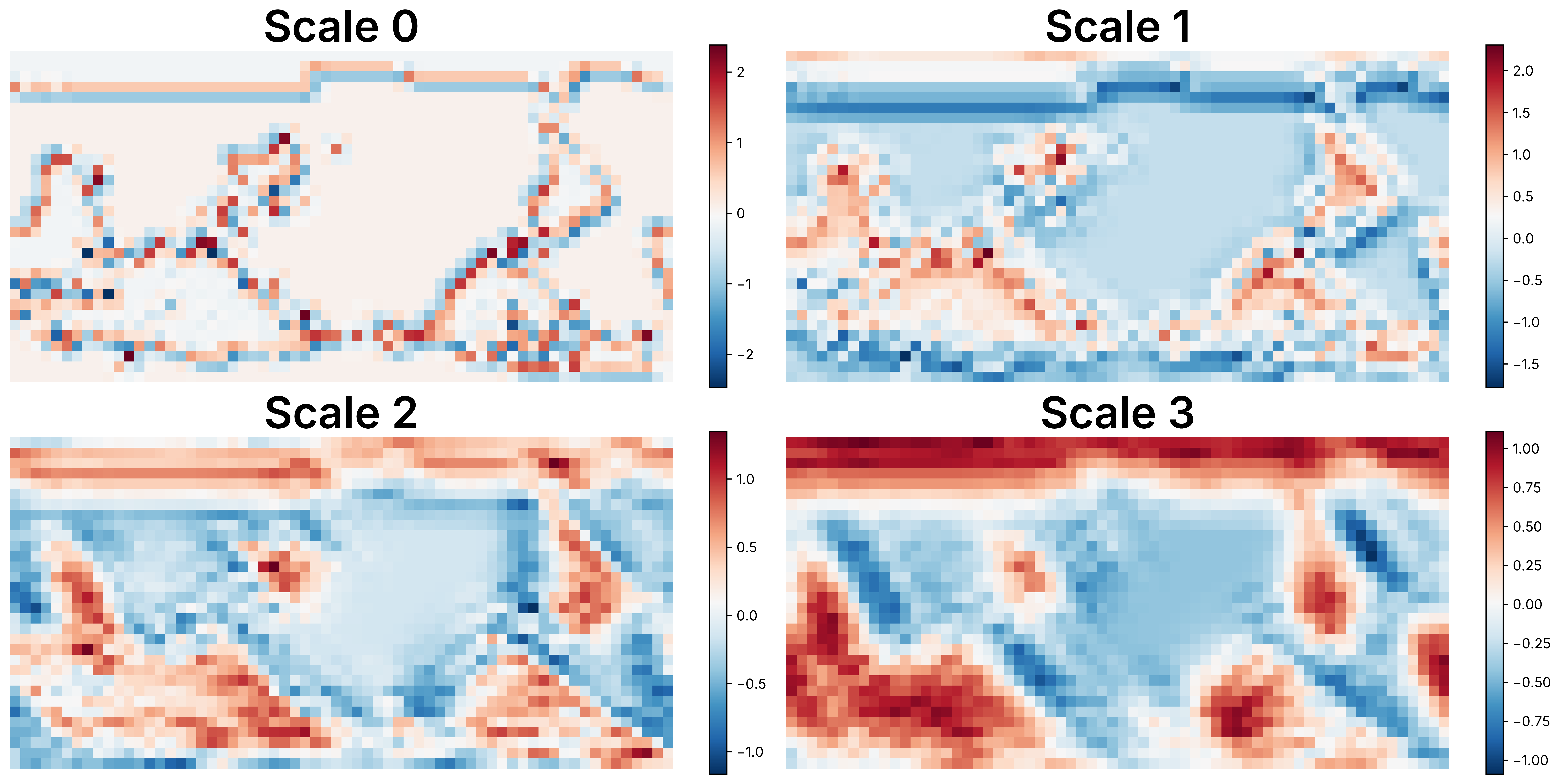} 
\caption{Visualization of scale-dependent spherical filters at multiple spatial scales.
}
\label{fig:visual_operators}
\end{figure}

We first quantitatively assess the impact of the multi-scale hierarchy within the spherical encoder. As shown in Table~\ref{tab:spatial_scales}, increasing the number of spatial scales from 1 (a non-multi-scale variant) to 4 leads to consistent improvements across all metrics. This trend demonstrates that explicit cross-scale modeling on the sphere is essential for capturing the complex dependencies of high-resolution atmospheric fields. While further increasing the scales to 8 yields marginal gains, the transition from single-scale to a 4-scale decomposition confirms that moderate hierarchical modeling is sufficient to enhance both predictive accuracy and numerical stability. We visualize Laplacian components at different levels to analyze multi-scale representations. Figure~\ref{fig:visual_operators} shows the operator sequence $G_s$ induces distinct hierarchical decomposition: Scale 0 captures localized high-frequency residuals, and Scale 3 isolates low-frequency large-scale structural trends. Analysis of adaptive modulation $\tilde{\mathbf{z}}_s$ shows MSPNO dynamically prioritizes spatial frequencies by local complexity: fine-scale components are emphasized in regions with active meteorological transitions, and global features dominate stable areas. Detailed visualizations are provided in Appendix~E.

\subsection{Evaluation of Physical Consistency}
We evaluate the physical fidelity of downscaled fields by calculating the residuals of atmospheric governing laws. As shown in Table \ref{tab:physical_consistency}, MSPNO yields the lowest residuals across all metrics, achieving a 6.5\% reduction in total physical violation compared to the strongest baseline. 
This advantage confirms that the synergy between spherical laplacian decomposition and differentiable constraints effectively enforces dynamical consistency, ensuring that MSPNO produces physically plausible regional climate projections.

\subsection{Ablation Studies}

\paragraph{Role of Explicit Spherical Encoding.}
Substituting spherical convolutions with standard 2D kernels (\textit{w/ Encoder (repl. Sphere)}) leads to consistent performance degradation (Table~\ref{tab:ablation_kernel_physics}). 
This degradation confirms that standard kernels, which assume a flat Euclidean grid, fail to adapt to the varying spatial densities of the spherical discretization.
Consequently, by integrating Spherical Laplacian decomposition, our approach faithfully captures the hierarchical nature of dynamics.

\paragraph{Merits of Localized Spherical Integration Operator.}
To assess the impact of spherical locality, we replace the Localized Spherical Integral Operator with a global interaction mechanism (\textit{w/ Attn. Kernel (repl. Sphere)}). This results in a significant increase in averaged RMSE. The performance gap highlights that localized spherical interactions are superior for preserving scale-consistent atmospheric structures and enabling information aggregation compared to global operators.

\paragraph{Impact of Physics-Guided Objective.} 
Discarding the physics-guided loss (\textit{w/o Physics-Guided Objective}) degrades all metrics. These results suggest that the physics-informed constraints enhance the numerical stability and structural fidelity of the learned operator, thereby promoting more physically plausible downscaling predictions.

\subsection{Robustness to Temporal Training Resolution}
Training across 3, 6, and 9-hour intervals reveals remarkably consistent performance (with the 6-hour interval being marginally optimal), proving that MSPNO generalizes effectively without requiring high-frequency temporal supervision. Detailed experimental table is shown in Appendix~F.

\section{Conclusion}
In this paper, we propose an operator learning framework called the \textbf{Multi-scale Spherical Physics-informed Neural Operator (MSPNO)} for meteorological downscaling. To address the challenges of multi-scale atmospheric dynamics and spherical geometric distortions, we devise two core components: the Spherical Laplacian Decomposition and the localized spherical integral operator. We then conducted comprehensive experiments on representative meteorological methods across varied variables and downscaling factors. The superior performance compared with baselines verified the effectiveness and robustness of our method. This work represents an attempt to integrate explicit spherical manifold geometry and physical constraints into neural operators, and it suggests a promising direction for large-scale neural surrogate models of Earth system processes and related scientific applications.

\clearpage
\section*{Acknowledgments}

This work is partially supported by the National Natural Science Foundation of China (No. 62372336), the Key Research and Development Program for Technological Innovation Project of Hubei Province (No. 2025BAB020).
Yiqiang Ye and Yichi Wang contributed equally. Chunxia Xiao and Haodi Zhang are the corresponding authors.

\bibliographystyle{named}
\bibliography{ijcai26}

\end{document}